\documentclass[11pt,letterpaper]{article}
\usepackage{emnlp2017}
\usepackage{times}
\usepackage{latexsym}
\usepackage{latexsym}
\usepackage{CJK}
\usepackage{graphicx}
\usepackage{multirow}
\usepackage{url}

\emnlpfinalcopy

\def\emnlppaperid{619}

\title{ Shallow Discourse Parsing with Maximum Entropy Model }

\author{Jingjing Xu \\
MOE Key Laboratory of Computational Linguistics, Peking University\\
School of Electronics Engineering and Computer Science, Peking University\\
{jingjingxu}@pku.edu.cn}

\date{}

\begin{document}

\maketitle

\begin{abstract}

In recent years, more research has been devoted to studying the subtask of the complete shallow discourse parsing, such as indentifying discourse connective and arguments of connective. There is a need to design a full discourse parser to pull these subtasks together. So we develop a discourse parser turning the free text into discourse relations. The parser includes connective identifier, arguments identifier, sense classifier and non-explicit identifier, which connects with each other in pipeline. Each component applies the maximum entropy model with abundant lexical and syntax features extracted from the Penn Discourse Tree-bank. The head-based representation of the PDTB is adopted in the arguments identifier, which turns the problem of indentifying the arguments of discourse connective into finding the head and end of the arguments. In the non-explicit identifier, the contextual type features like words which have high frequency and can reflect the discourse relation are introduced to improve the performance of non-explicit identifier. Compared with other methods, experimental results achieve the considerable performance.

\end{abstract}

\section{Introduction}
\label{introduction}

Automated deriving discourse relation from free text is a challenging but im-portant problem. The shallow discourse parsing is very useful in the text summariza-tion~\cite{Marcu00rhetorical}, opinion analysis~\cite{SomasundaranNWG09} and natural language generation. 
Shallow discourse parser is the system of parsing raw text into a set of discourse relations between two adjacent or non-adjacent text spans. Discourse relation is composed of a discourse connective, two arguments of the discourse connective and the sense of the discourse connective. Discourse connective signals the explicit dis-course relation, but in non-explicit discourse relation, a discourse connective is omit-ted. Two arguments of the discourse connective, Arg1 and Arg2, which are the two adjacent or non-adjacent text spans connecting in the discourse relation. The sense of the discourse connective characterizes the nature of the discourse relations. 
The following discourse relation annotation is taken from the document in the PDTB. Arg1 is shown in italicized, and Arg2 is shown in bold. The discourse connective is underlined.

The connective identifier finds the connective word, “unless”. The arguments identifier locates the two arguments of “unless”. The sense classifier labels the dis-course relation. 
The non-explicit identifier checks all the pair of adjacent sentences. If the non-explicit identifier indentifies the pair of sentences as non-explicit relation, it will label it the relation sense.
Though many research work~\cite{WellnerPHRS06,PitlerRMNLJ08,PitlerLN09} are committed to the shallow discourse parsing field, all of them are focus on the subtask of parsing only rather than the whole parsing process. Given all that, a full shallow discourse parser framework is proposed in our paper to turn the free text into discourse relations set. The parser includes connective identifier, arguments identifier, sense classifier and non-explicit identifier, which connects with each other in pipeline.
In order to enhance the performance of the parser, the feature-based maximum entropy model approach is adopted in the experiment. Maximum entropy model offers a clean way to combine diverse pieces of contextual evidence in order to estimate the probability of a certain linguistic class occurring with a certain linguistic context in a simple and accessible manner. 
The three main contributions of the paper are:

\begin{itemize}
\item  The subtasks of the shallow discourse parsing are pulled together in pipeline. 

\item Two different approaches are proposed to evaluate the performance of the parser with maximum entropy model.
 
\item  The contextual type features indicating the non-explicit discourse relation are in-troduced in non-explicit identifier, like words which have a high frequency occur-rence in Arg2.

\end{itemize}
The rest of this paper is organized as follows. Section 2 reviews related work in discourse parsing. Section 3 describes the experimental corpus--PDTB. Section 4 de-scribes the framework and the components of the parser. Section 5 presents experi-ments and evaluations. Conclusions are presented in the Section 6.

\section{Related Work}

Different from traditional shallow parsing~\cite{SunMOTT08,SunMOT09,SunLWL14} which is dealing with a single sentence, the shallow discourse parsing tries to analyze the discourse level information, which is more complicated.
Since the release of second version of the Penn Discourse Treebank (PDTB), which is over the 1 million word Wall Street Journal corpus, analyzing the PDTB-2.0 is very useful for further study on shallow discourse parsing. Prasad et al.~\shortcite{PrasadDLMRJW08} describe lexically-grounded annotations of discourse relations in PDTB. 
Identifying the discourse connective from ordinary words accurately is not easy because discourse words can have discourse or non-discourse usage. Pitler and Nenkova~\shortcite{PitlerN09} use syntax feature to disambiguate explicit discourse connective in text and prove that the syntactic features can improve performance in disambiguation task.
After identifying the discourse connective, there is a need to find the arguments. There are some different methods to find the arguments. Ziheng Lin et al.~\shortcite{LinNK14} first identify the locations of Arg1, and choose sentence from prior candidate sentence if the location is before the connective. Otherwise, label arguments span by choosing the high node in the parse tree. Wellner and Pustejovsky~\shortcite{WellnerP07} focus on identifying rela-tions between the pairs of head words. Based on such thinking, Robert Elwell and Jason Baldridge~\shortcite{ElwellB08} improve the performance using connective specific rankers, which differentiate between specific connectives and types of connectives.
Ziheng Lin et al.~\shortcite{LinNK14} present an implicit discourse relation classifier based the Penn Discourse Treebank.
All of these efforts can be viewed as the part of the full parser. More and more researcher has been devoted to the subtask of the shallow discourse parsing, like dis-ambiguating discourse connective~\cite{PitlerN09}, finding implicit relation~\cite{LinNK14}. There is a need to pull these subtasks together to achieve more efforts. So in this paper, we develop a full shallow discourse parser based on the maximum entropy model using abundant features. Our parser attempts to identify connective, arguments of discourse connec-tive and the relation into right sense.

\begin{table*}[!hbt]
\centering
\caption{Features on connective identifier.}
\label{my-label}
\begin{tabular}{|l|l|}
\hline
Feature                & Description                                                                                                     \\ \hline
Self Category          & For single word connectives, this is the POS tag of the word\\ \hline
Parent Category        & Self Category’s immediate parent                                                                                \\ \hline
Left Sibling Category  & The immediate left sibling of the Self Category, if not, this is “NONE”                                         \\ \hline
Right Sibling Category & The immediate right sibling of the Self Category, if not, this is “NONE”                                        \\ \hline
Word                   & Connective word                                                                                                 \\ \hline
RightVp                & Whether the right sibling category contains a VP or not                                                         \\ \hline
Prev POS               & POS of the word immediately before Connective word                                                              \\ \hline
Prev POS+word POS      & Prev POS\&word POS                                                                                              \\ \hline
Next POS               & POS of the word immediately after connective word                                                               \\ \hline
Next POS+word POS      & Next POS\&word POS                                                                                              \\ \hline
\end{tabular}
\end{table*}

\begin{table*}[!hbt]
\centering
\caption{Features on arguments identifier.}
\label{my-label}
\begin{tabular}{|l|l|}
\hline
Feature index & Description                                                       \\ \hline
1             & Whether the candidate is in the same sentence with the connective \\ \hline
2             & Connective word                                                   \\ \hline
3             & Down-case connective word                                         \\ \hline
4             & Candidate word                                                    \\ \hline
5             & Before or after the connective word                               \\ \hline
6             & Connective where in the sentence (beginning, middle, end)         \\ \hline
7             & 2,6                                                               \\ \hline
8             & Head to connective through the constituent tree                   \\ \hline
9             & Collapsed path without part-of-speech                             \\ \hline
10            & The length of 8                                                   \\ \hline
11            & Dependency path from argument to connective                       \\ \hline
12            & Type of connective                                                \\ \hline
\end{tabular}
\end{table*}

\begin{table}[!hbt]
\centering
\caption{ The pseudocode of arguments identifier.}
\label{my-label}
\begin{tabular}{|ll|}
\hline

1                                                   & Input: connective word x and raw text T \\ \hline
2                                                   & Initialize Score1=\{\}, Score2=\{\}     \\ \hline
3                                                   & Get the Arg1 candidate set C1 of x      \\ \hline
4                                                   & for c1 in C1                            \\ \hline
5                                                   & s1=rank(c1)                             \\ \hline
6                                                   & Score1=Score1                           \\ \hline
7                                                   & end for                                 \\ \hline
8                                                   & Get the Arg2 candidate set C2 of x      \\ \hline
9                                                   & for c2 in C2                            \\ \hline
10                                                  & s2=rank(c2)                             \\ \hline
11                                                  & Score2=Score2                           \\ \hline
12                                                  & end for                                 \\ \hline
13                                                  & Find c1, let                            \\ \hline
14                                                  & Find nature end of Arg1                 \\ \hline  
15                                                  & Get Arg1                                \\ \hline
16                                                  & Find c2, let                            \\ \hline
17                                                  & Find nature end of Arg2                 \\ \hline
18                                                  & Get Arg2                                \\ \hline
19                                                  & return Arg1, Arg2                       \\ \hline
\end{tabular}
\end{table}

\begin{table*}
\centering
\caption{The overall performance over explicit relation and non-explicit relation.}
\label{my-label}
\begin{tabular}{|l|l|l|l|l|}
\hline
         &     &Precision  & Recall      & F1-Measure   \\ \hline
Connective & Our parser  & 92.60\%    & 90.00\% & 91.30\% \\ \hline
           & Baseline\_1 & 73.50\%    & 39.20\% & 51.10\% \\ \hline
           & Baseline\_2 & 78.70\%    & 71.20\% & 74.80\% \\ \hline
Arg1       & Our parser  & 48.50\%    & 47.10\% & 47.80\% \\ \hline
           & Baseline\_1 & 23.00\%    & 18.60\% & 20.60\% \\ \hline
           & Baseline\_2 & 36.60\%    & 35.60\% & 36.10\% \\ \hline
Arg2       & Our parser  & 68.90\%    & 71.40\% & 70.10\% \\ \hline
           & Baseline\_1 & 61.70\%    & 50.40\% & 55.50\% \\ \hline
           & Baseline\_2 & 61.20\%    & 59.40\% & 60.30\% \\ \hline
Sense      & Our parser  & 68.00\%    & 68.30\% & 68.10\% \\ \hline
           & Baseline\_1 & 20.50\%    & 20.10\% & 20.30\% \\ \hline
           & Baseline\_2 & 58.70\%    & 57.00\% & 57.80\% \\ \hline
\end{tabular}
\end{table*}

\section{The Penn Discourse Treebank}

The Penn Discourse Treebank is the corpus which is over one million words from the Wall Street Journal~\cite{treebank} , annotated with discourse relations. The table one shows the discourse relation extracted from PDTB. Arg1 is shown in italicized, Arg2 is shown in bold. The discourse connective is underlined.

Discourse connective is the signal of explicit relation. Discourse connective in the PTDB can be classified as three categories: subordinating conjunctions (e.g., because, if, etc.), coordinating conjunctions (e.g., and, but, etc.), and discourse adverbials (e.g., however, also, tec.). Different category has different discourse usage. 
Discourse connective word can be ambiguous between discourse or non-discourse usage. An apparent example is 'after' because it can be a VP (e.g., "If you are after something, you are trying to get it") or it can be a connective (e.g., “It wasn't until after Christmas that I met Paul”). 
In the case of explicit relation, Arg2 is the argument to which the connective is syntactically bound, and Arg1 is the other argument. But the span of the arguments of explicit relation can be clauses or sentences. In the case of implicit relation, Arg1 is before Arg2~\cite{PrasadDLMRJW08}.
For explicit, implicit and altLex relation, there are three-level hierarchy of relation senses. The first level consists of four major relation classes: Temporal, Contingency, Comparison, and Expansion.

\section{Shallow Discourse Parser framework}

We design a complete discourse parser connecting subtasks together in pipeline. First let’s have a quick view about the procedure of the parser. The first step is pre-processing, which takes the raw text as input and generates POS tag of token, the dependency tree, constituent tree and so on. Next the parser needs to distinguish the connective between discourse usage and non-discourse usage. Then, the two argu-ments of discourse connective need to be identified. Next to above steps, the parser labels the discourse relation right sense. Until now the explicit relations already have been found fully. The last step is indentifying the non-explicit relation. The parser will handle every pair of adjacent sentences in same paragraph. 
The text is pre-processed by the Stanford CoreNLP tools. Stanford CoreNLP provides a series of natural language analysis tools which can tokenize the text, label tokens with their part-of-speech (POS) tag, and provides full syntactic analysis, in-cluding both constituent and dependency representation. The parser uses Stanford CoreNLP toolkit to preprocess the raw text. Next, each component of the parser will be described in detail.

\subsection{	Connective Identifier}

The main duty of this component is disambiguate the connective words which are in PDTB predefined set. Pitler and Nenkova~cite{PitlerN09} show that syntactic features are very useful on disambiguate discourse connective, so we adopt these syntactic fea-tures as part of our features. Ziheng Lin et al.~\shortcite{LinKN09} show that a connective’s context and part-of-speech (POS) gives a very strong indication of discourse usage. The table 1 shows the feature we use.

\subsection{Arguments Identifier}

On this step, we adopt the head-based thinking~\cite{WellnerP07}, which turns the problem of identifying arguments of discourse connective into identifying the head and end of the arguments. First, we need to extract the candidates of arguments. To reduce the Arg1 candidates space, we only consider words with appropriate part-of-speech (all verbs, common nouns, adjectives) and within 10 ”steps” between word and connec-tive as candidates, where a step is either a sentence boundary or a dependency link. Only words in the same sentence with the connective are considered for Arg2 candi-dates. Second, we need to choose the best candidate as the head of Arg1 and Arg2. In the end, we need to obtain the arguments span according head and end of argu-ments on the constituent tree. The table 2 shows the feature we use. The table 3 shows the procedure of the arguments identifier.

\subsection{Sense Classifier}

The sense of discourse relation has three levels: class, type and subtype.
There are four classes on the top level of the sense: Comparison, Temporal , Con-tingency, Expansion. Each class includes a set of different types, and some types may have different subtypes.
The connective itself is a very good feature because discourse connective almost determine senses. So we train an explicit classifier using simple but effective features.

\subsection{Non-explicit Identifier}

The non-explicit relation is the relation between adjacent sentences in same para-graph. So we just check adjacent sentences which don’t form explicit relation and then label them with non-explicit relation or nothing. In the experiment, we find that the two arguments of non-explicit relation have association with each other and also have some common words. So we introduce feature words, which indicate appear-ance of relation, like “it, them”.

\section{Experiments}

In our experiments, we make use of the Section 02-21 in the PDTB as training set, Section 22 as testing set. All of components adopt maximum entropy model. 
In order to evaluate the performance of the discourse parser, we compare it with other approaches: (1) Baseline\_1, which applies the probability information. The connective identifier predicts the connective according the frequency of the connec-tive in the train set. The arguments identifier takes the immediately previous sentence in which the connective appears as Arg1 and the text span after the connective but in the same sentence with connective as Arg2. The non-explicit identifier labels the ad-jacent sentences according to the frequency of the non-explicit relation. (2) Base-line\_2, which is the parser using the Support Vector Maching as the train and predic-tion model with numeric type feature from the hashcode of the textual type feature.

It is not surprised to find that Baseline\_1 shows the poorest performance, which it just considers the probability information, ignores the contextual link. The perfor-mance of Baseline\_2 is better than that of “Baseline\_1”. This can be mainly credited to the ability of abundant lexical and syntax features. Our parser shows better per-formance than Baselin\_2 because the most of features we use are textual type fea-tures, which are convenient for the maximum entropy model. Though the textual type features can turn into numeric type according to hashcode of string, it is incon-venient for Support Vector Machine because the hashcode of string is not continu-ous. 
According the performance of the parser, we find that the connective identifying can achieve higher precision and recall rate. In addition, the precision and recall rate of identifying Arg2 is higher than that of identifying Arg1 because Arg2 has stronger syntax link with connective compared to Arg1. The sense has three layers: class, type and subtype.

\section{Conclusion}

In this paper, we design a full discourse parser to turn any free English text into discourse relation set. The parser pulls a set of subtasks together in a pipeline. On each component, we adopt the maximum entropy model with abundant lexical, syntactic features. In the non-explicit identifier, we introduce some contextual infor-mation like words which have high frequency and can reflect the discourse relation to improve the performance of non-explicit identifier.
In addition, we report another two baselines in this paper, namely Baseline\-1 and Baseline\-2, which base on probabilistic model and support vector machine model, respectively. Compared with two baselines, our parser achieves the considerable improvement. As future work, we try to explore the deep learning methods~\cite{XuS16,SunRMW17,MA2017,HeS17,XuS17,shumingma,iclr2018} to improve this study.
We believe that our discourse parser is very useful in many applications because we can provide the full discourse parser turning any unrestricted text into discourse structure.

\bibliography{emnlp2017}

\begin{thebibliography}{}
\expandafter\ifx\csname natexlab\endcsname\relax\def\natexlab#1{#1}\fi

\bibitem[{Elwell and Baldridge(2008)}]{ElwellB08}
Robert Elwell and Jason Baldridge. 2008.
\newblock Discourse connective argument identification with connective specific
  rankers.
\newblock In {\em ICSC\/}. IEEE Computer Society, pages 198--205.

\bibitem[{He and Sun(2017)}]{HeS17}
Hangfeng He and Xu~Sun. 2017.
\newblock A unified model for cross-domain and semi-supervised named entity
  recognition in chinese social media.
\newblock In {\em Proceedings of the Thirty-First {AAAI} Conference on
  Artificial Intelligence, February 4-9, 2017, San Francisco, California,
  {USA.}\/}. pages 3216--3222.

\bibitem[{Lin et~al.(2009)Lin, Kan, and Ng}]{LinKN09}
Ziheng Lin, Min-Yen Kan, and Hwee~Tou Ng. 2009.
\newblock Recognizing implicit discourse relations in the penn discourse
  treebank.
\newblock In {\em EMNLP\/}. ACL, pages 343--351.

\bibitem[{Lin et~al.(2014)Lin, Ng, and Kan}]{LinNK14}
Ziheng Lin, Hwee~Tou Ng, and Min-Yen Kan. 2014.
\newblock A pdtb-styled end-to-end discourse parser.
\newblock {\em Natural Language Engineering\/} 20(2):151--184.

\bibitem[{Ma and Sun(2017)}]{shumingma}
Shuming Ma and Xu~Sun. 2017.
\newblock A semantic relevance based neural network for text summarization and
  text simplification.
\newblock {\em CoRR\/} abs/1710.02318.

\bibitem[{Ma et~al.(2017)Ma, Sun, Xu, Wang, Li, and Su}]{MA2017}
Shuming Ma, Xu~Sun, Jingjing Xu, Houfeng Wang, Wenjie Li, and Qi~Su. 2017.
\newblock Improving semantic relevance for sequence-to-sequence learning of
  chinese social media text summarization.
\newblock In {\em ACL'17\/}.

\bibitem[{Marcu(2000)}]{Marcu00rhetorical}
Daniel Marcu. 2000.
\newblock {The rhetorical parsing of unrestricted texts: a surface-based
  approach}.
\newblock {\em Comput. Linguist.\/} 26(3):395--448.

\bibitem[{Marcus et~al.(1993)Marcus, Santorini, and Marcinkiewicz}]{treebank}
Mitchell~P. Marcus, Beatrice Santorini, and Mary~Ann Marcinkiewicz. 1993.
\newblock Building a large annotated corpus of {E}nglish: {T}he {P}enn
  treebank.
\newblock {\em Computational Linguistics\/} 19(2):313--330.

\bibitem[{Pitler et~al.(2009)Pitler, Louis, and Nenkova}]{PitlerLN09}
Emily Pitler, Annie Louis, and Ani Nenkova. 2009.
\newblock Automatic sense prediction for implicit discourse relations in text.
\newblock In {\em ACL/IJCNLP\/}. The Association for Computer Linguistics,
  pages 683--691.

\bibitem[{Pitler and Nenkova(2009)}]{PitlerN09}
Emily Pitler and Ani Nenkova. 2009.
\newblock Using syntax to disambiguate explicit discourse connectives in text.
\newblock In {\em ACL/IJCNLP (Short Papers)\/}. The Association for Computer
  Linguistics, pages 13--16.

\bibitem[{Pitler et~al.(2008)Pitler, Raghupathy, Mehta, Nenkova, Lee, and
  Joshi}]{PitlerRMNLJ08}
Emily Pitler, Mridhula Raghupathy, Hena Mehta, Ani Nenkova, Alan Lee, and
  Aravind~K. Joshi. 2008.
\newblock Easily identifiable discourse relations.
\newblock In {\em COLING (Posters)\/}. pages 87--90.

\bibitem[{Prasad et~al.(2008)Prasad, Dinesh, Lee, Miltsakaki, Robaldo, Joshi,
  and Webber}]{PrasadDLMRJW08}
Rashmi Prasad, Nikhil Dinesh, Alan Lee, Eleni Miltsakaki, Livio Robaldo,
  Aravind~K. Joshi, and Bonnie~L. Webber. 2008.
\newblock The penn discourse treebank 2.0.
\newblock In {\em LREC\/}. European Language Resources Association.

\bibitem[{Somasundaran et~al.(2009)Somasundaran, Namata, Wiebe, and
  Getoor}]{SomasundaranNWG09}
Swapna Somasundaran, Galileo Namata, Janyce Wiebe, and Lise Getoor. 2009.
\newblock Supervised and unsupervised methods in employing discourse relations
  for improving opinion polarity classification.
\newblock In {\em EMNLP\/}. ACL, pages 170--179.

\bibitem[{Sun et~al.(2014)Sun, Li, Wang, and Lu}]{SunLWL14}
Xu~Sun, Wenjie Li, Houfeng Wang, and Qin Lu. 2014.
\newblock Feature-frequency-adaptive on-line training for fast and accurate
  natural language processing.
\newblock {\em Computational Linguistics\/} 40(3):563--586.

\bibitem[{Sun et~al.(2009)Sun, Matsuzaki, Okanohara, and Tsujii}]{SunMOT09}
Xu~Sun, Takuya Matsuzaki, Daisuke Okanohara, and Jun'ichi Tsujii. 2009.
\newblock Latent variable perceptron algorithm for structured classification.
\newblock In {\em Proceedings of the 21st International Joint Conference on
  Artificial Intelligence (IJCAI 2009)\/}. pages 1236--1242.

\bibitem[{Sun et~al.(2008)Sun, Morency, Okanohara, and Tsujii}]{SunMOTT08}
Xu~Sun, Louis-Philippe Morency, Daisuke Okanohara, and Jun'ichi Tsujii. 2008.
\newblock Modeling latent-dynamic in shallow parsing: A latent conditional
  model with improved inference.
\newblock In {\em COLING\/}. pages 841--848.

\bibitem[{Sun et~al.(2017{\natexlab{a}})Sun, Ren, Ma, and Wang}]{SunRMW17}
Xu~Sun, Xuancheng Ren, Shuming Ma, and Houfeng Wang. 2017{\natexlab{a}}.
\newblock meprop: Sparsified back propagation for accelerated deep learning
  with reduced overfitting.
\newblock In {\em Proceedings of the 34th International Conference on Machine
  Learning, {ICML} 2017, Sydney, NSW, Australia, 6-11 August 2017\/}. pages
  3299--3308.

\bibitem[{Sun et~al.(2017{\natexlab{b}})Sun, Wei, Ren, and Ma}]{iclr2018}
Xu~Sun, Bingzhen Wei, Xuancheng Ren, and Shuming Ma. 2017{\natexlab{b}}.
\newblock Label embedding network: Learning label representation for soft
  training of deep networks.
\newblock {\em CoRR\/} abs/1710.10393.

\bibitem[{Wellner and Pustejovsky(2007)}]{WellnerP07}
Ben Wellner and James Pustejovsky. 2007.
\newblock Automatically identifying the arguments of discourse connectives.
\newblock In {\em EMNLP-CoNLL\/}. ACL, pages 92--101.

\bibitem[{Wellner et~al.(2006)Wellner, Pustejovsky, Havasi, Rumshisky, and
  Saurí}]{WellnerPHRS06}
Ben Wellner, James Pustejovsky, Catherine Havasi, Anna Rumshisky, and Roser
  Saurí. 2006.
\newblock Classification of discourse coherence relations: An exploratory study
  using multiple knowledge sources.
\newblock In {\em SIGDIAL Workshop\/}. The Association for Computer
  Linguistics, pages 117--125.

\bibitem[{Xu et~al.(2017)Xu, Ma, Zhang, Wei, Cai, and Sun}]{XuS17}
Jingjing Xu, Shuming Ma, Yi~Zhang, Bingzhen Wei, Xiaoyan Cai, and Xu~Sun. 2017.
\newblock Transfer learning for low-resource chinese word segmentation with a
  novel neural network.
\newblock In {\em The Conference on Natural Language Processing and Chinese
  Computing\/}.

\bibitem[{Xu and Sun(2016)}]{XuS16}
Jingjing Xu and Xu~Sun. 2016.
\newblock Dependency-based gated recursive neural network for chinese word
  segmentation.
\newblock In {\em ACL'16\/}. pages 567--572.

\end{thebibliography}

\bibliographystyle{emnlp_natbib}

\end{document}